\documentclass[letterpaper]{article} 
\usepackage{aaai25}  
\usepackage{times}  
\usepackage{helvet}  
\usepackage{courier}  
\usepackage[hyphens]{url}  
\usepackage{graphicx} 
\usepackage{natbib}  
\usepackage{caption} 
\usepackage{algorithm}
\usepackage{algorithmic}
\usepackage{tikz}
\def\checkmark{\tikz\fill[scale=0.4](0,.35) -- (.25,0) -- (1,.7) -- (.25,.15) -- cycle;} 
\usepackage{newfloat}
\usepackage{listings}
\DeclareCaptionStyle{ruled}{labelfont=normalfont,labelsep=colon,strut=off} 
\floatstyle{ruled}
\newfloat{listing}{tb}{lst}{}
\floatname{listing}{Listing}
\title{ICM-Assistant: Instruction-tuning Multimodal Large Language Models for Rule-based Explainable Image Content Moderation}
\author{
     Mengyang Wu\textsuperscript{\rm 1,\rm 2,\equalcontrib}\thanks{Internship with Huawei Hong Kong Research Center.}, Yuzhi Zhao\textsuperscript{\rm 2, \equalcontrib}\thanks{Corresponding Author and Project Lead.},
    Jialun Cao\textsuperscript{\rm 3}, Mingjie Xu\textsuperscript{\rm 4}, Zhongming Jiang\textsuperscript{\rm 4}, Xuehui Wang\textsuperscript{\rm 5}, Qinbin Li\textsuperscript{\rm 6}, Guangneng Hu\textsuperscript{\rm 7}, Shengchao Qin\textsuperscript{\rm 8,\rm 9}, Chi-Wing Fu\textsuperscript{\rm 1}
}
\affiliations{
    \textsuperscript{\rm 1}Department of Computer Science and Engineering, The Chinese University of Hong Kong
    
    \textsuperscript{\rm 2}Huawei Hong Kong Research Center
    
    \textsuperscript{\rm 3}Department of Computer Science and Engineering, The Hong Kong University of Science and Technology
    
    \textsuperscript{\rm 4}Huawei 2012 Laboratories
    
    \textsuperscript{\rm 5}Artificial Intelligence Institute, Shanghai Jiao Tong University
    
    \textsuperscript{\rm 6}School of Computer Science and Technology, Huazhong University of Science and Technology
    
    \textsuperscript{\rm 7}School of Computer Science and Technology, Xidian University
    
    \textsuperscript{\rm 8}Guangzhou Institute of Technology, Xidian University
    
    \textsuperscript{\rm 9}ICTT and ISN Laboratory, Xidian University

    {mywu}@cse.cuhk.edu.hk;
    {yzzhao2-c}@my.cityu.edu.hk
}

\usepackage{bibentry}

\begin{document}

\maketitle

\begin{abstract}
Controversial contents largely inundate the Internet, infringing various cultural norms and child protection standards. Traditional Image Content Moderation (ICM) models fall short in producing precise moderation decisions for diverse standards, while recent multimodal large language models (MLLMs), when adopted to general rule-based ICM, often produce classification and explanation results that are inconsistent with human moderators. Aiming at flexible, explainable, and accurate ICM, we design a novel rule-based dataset generation pipeline, decomposing concise human-defined rules and leveraging well-designed multi-stage prompts to enrich short explicit image annotations. Our ICM-Instruct dataset includes detailed moderation explanation and moderation Q-A pairs. Built upon it, we create our ICM-Assistant model in the framework of rule-based ICM, making it readily applicable in real practice. Our ICM-Assistant model demonstrates exceptional performance and flexibility. Specifically, it significantly outperforms existing approaches on various sources, improving both the moderation classification (36.8\% on average) and moderation explanation quality (26.6\% on average) consistently over existing MLLMs. 

\noindent{Caution: Content includes offensive language or images.}
\end{abstract}


\begin{links}
    \link{Code}{https://github.com/zhaoyuzhi/ICM-Assistant}
\end{links}

\begin{figure*}[ht!]
\centering
\includegraphics[width=.99\linewidth]{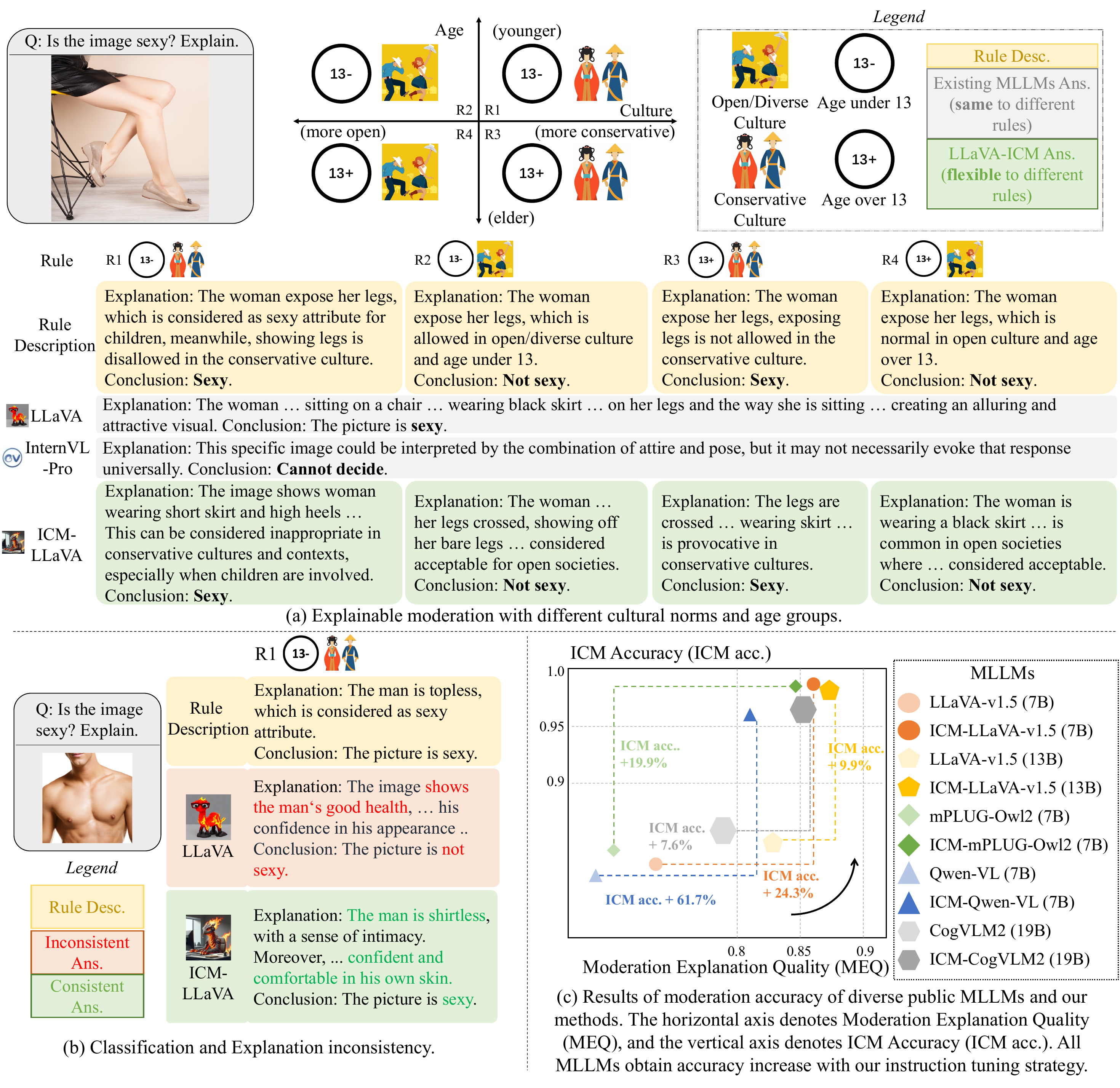}
\caption{Overall framework: With input image and rules (on cultural norm and children protection), our method can (a) \textbf{flexibly} align with human moderators with four rules, and provide \textbf{explainable} results, overcoming the (b) classification and explanation inconsistency and achieving (c) more \textbf{accurate} moderation classification and explanations than the baseline MLLMs.}
\label{fig:fig-teaser}
\end{figure*}

\section{Introduction}
\label{sec:intro}
In recent years, user-generated and AI-generated image contents (e.g., by DALL-E~\cite{ramesh2021zero} or stable diffusion~\cite{rombach2022high}) keep multiplying on online platforms. Yet, some contents are against platform rules on child protection (e.g., sexy, horrifying)~\cite{Content_policies_for_Facebook, Content_policies_for_Google_Search, Content_policies_for_TikTok} or on cultural regulations~\cite{de2023inequalities}. Therefore, rule-based image content moderation has become increasingly significant. 

Today, image content moderation faces three main challenges. First, moderation rules vary with cultures and ages. Different moderation strategies are needed for individual cultural regulations or child protection (see Fig.~\ref{fig:fig-teaser} (a), different combinations of culture and age have different moderation results on the same image). Second, transparency and openness receive increasing attention~\cite{juneja2020through}. So, for the disclosure of review results, we need to provide reasons based on moderation rules, which increase the moderation complexity. Third, we need an in-depth understanding of the image contents to ensure accurate moderation results. Summarizing the challenges, there is a pressing need for {\em flexible\/}, {\em explainable\/} and {\em accurate\/} image content moderation (ICM) systems.

Traditional ICM systems~\cite{son2023reliable, lees2022new, akyon2023state, momo2023evaluation} and datasets~\cite{karavarsamis2013detecting, connie2017smart, NudeNet, phan2022lspd} typically formulate ICM as a classification task. However, the image violation clues are often obscure and not identifiable solely by specific patterns or scenes. Therefore, traditional methods often obtain low classification accuracy. Moreover, their moderation processes are not explainable and the datasets are not flexible to different rules.

Recent multimodal large language models (MLLMs)~\cite{liu2024visual, bai2023qwen, team2023gemini, achiam2023gpt, markov2023holistic} also perform ICM, by providing a general ability in describing and reasoning image contents. After practice, we find that existing MLLMs are not aligned with any specific moderation rules, resulting in two main inconsistencies with human moderators following various rules. (1) {\em Classification inconsistency\/}: the classification result/decision is inconsistent with the rules, and (2) {\em Explanation inconsistency\/}: the decision follows the rules but the explanation is wrong or incomplete, as shown in Fig.~\ref{fig:fig-teaser} (b). To address the issues, one solution is to instruction-tune MLLMs using domain-specific datasets. Yet, the lack of a multi-modal moderation dataset and the difficulty in customizing existing datasets into specific rules greatly limit the effectiveness and flexibility of the approach. To bridge the gap, we need an effective ICM dataset and an efficient data generation pipeline, such that we can customize and enhance the moderation classification and explanation of MLLMs, and adopt it in the framework of rule-based ICM.
\begin{figure*}[t!]
\centering
\includegraphics[width=1.\linewidth]{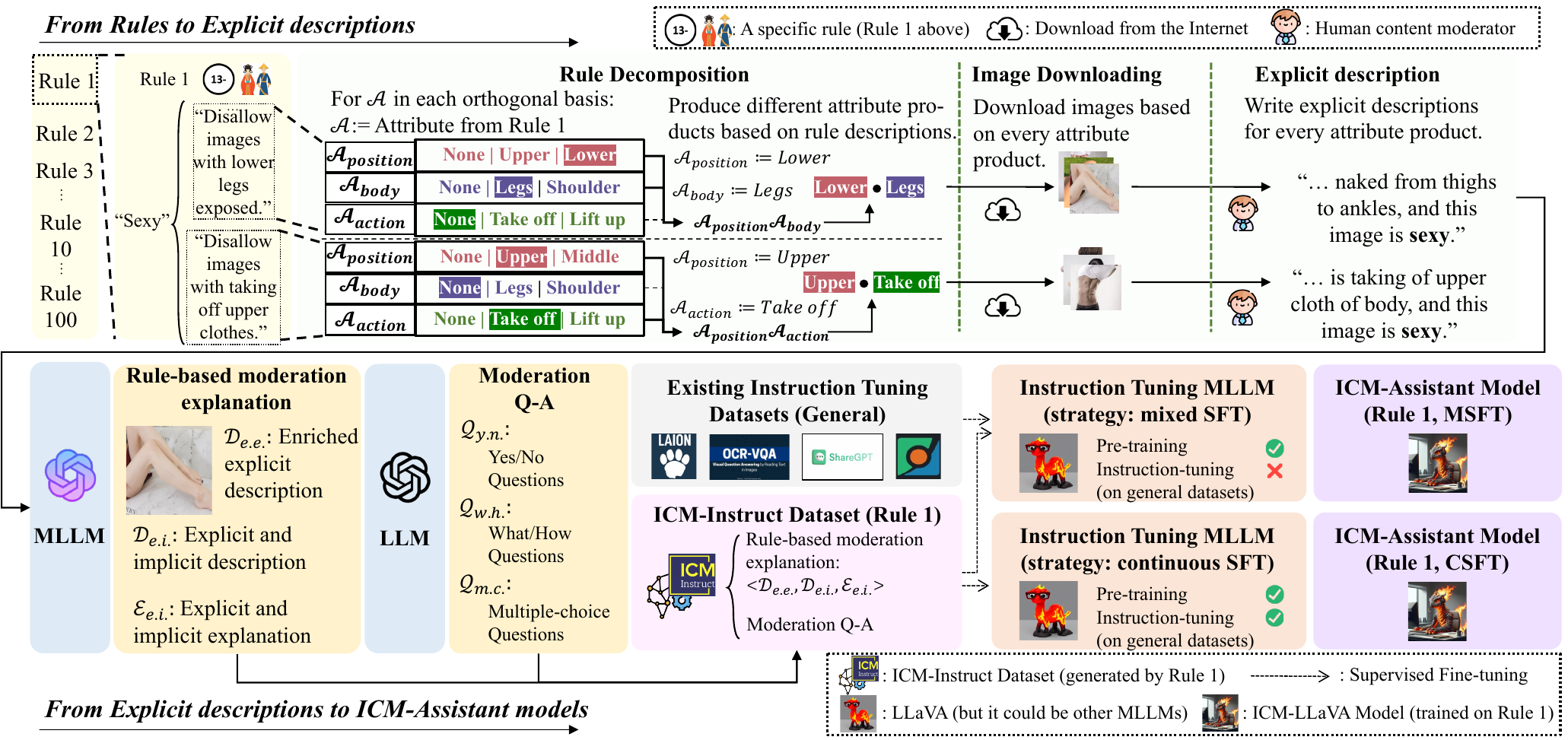}
\caption{The overall pipeline from a specific set of moderation rules to a rule-based ICM-Assistant model. The first row illustrates rule decomposition, image downloading, and initialization of explicit descriptions. The second row presents the fully automatic data augmentation pipeline for ICM-Instruct dataset and the instruction-tuning process for ICM-Assistant models.}
\label{fig:fig-framework}
\end{figure*}
We approach the goal by first designing a flexible rule-based ICM data generation pipeline. Given pre-defined moderation rules, we decompose it into several sub-categories, termed {\em attribute products\/}. Therefore, we can easily adapt to different moderation rules by changing the classification labels of attribute products. Then, we adopt {\em multi-stage prompts\/} to progressively generate and enrich moderation explanations and Q-A pairs from the short annotations based on the ``Chain-of-Thought'' (CoT)~\cite{wei2022chain}, which decomposes ICM to several steps and makes ICM more accurately and interpretably. Doing so aligns with the ICM task, which requires the production of a moderation explanation as part of the moderation result. Accordingly, we construct a large-scale instruction-tuning \textbf{ICM-Instruct dataset}. With our ICM-Instruct dataset, we thereby perform instruction-tuning on various MLLMs by injecting content moderation reasoning knowledge. Results show the strong capabilities of our model and dataset in achieving flexible, explainable, and accurate moderation, as shown in Fig.~\ref{fig:fig-teaser} (c).

There are three technical contributions in this work:

(1) The flexible dataset generation pipeline based on concise moderation rules, short human annotations, and well-designed multi-stage prompts for prompt-guided data augmentation, largely surpassing existing moderation datasets;

(2) A large-scale instruction-tuning ICM-Instruct dataset and an efficient evaluation benchmark for both the proposed benchmark and existing image moderation test set;

(3) Built upon the ICM-Instruct dataset, we produce the first method to accomplish diverse content moderation tasks, including moderation classification, explanation, and question-answering by adopting several instruction-tuning strategies to upgrade several MLLMs with ICM-Assistant.

Results demonstrate that our method well moderates a wide range of images, outperforming existing MLLMs by an average of 33.7\% in moderation classification and 26.6\% in moderation explanation on various moderation tasks, rules, and data sources (e.g., user-generated, AI-generated, etc.).

\section{Related Works}

\noindent \textbf{ICM datasets and methods.} \
Image content moderation is an important topic of great practical need, given the proliferation of online platforms and social media. Common datasets include nudity/sexual content~\cite{karavarsamis2013detecting, connie2017smart, NudeNet, phan2022lspd}, terrorism/violence~\cite{schedi2015vsd2014, roy2017automated, bianculli2020dataset}, etc. Method-wise, ICM can be single-modal~\cite{ son2023reliable, lees2022new, akyon2023state, momo2023evaluation, zeng2024shieldgemmagenerativeaicontent} or multimodal~\cite{gupta2018empowering, tang2021videomoderator, yuan2024rethinking}. While Large language models (LLMs) offer great help in data generation~\cite{markov2023holistic, jin2024gpt} and data augmentation~\cite{ma2023adapting}, the resulting datasets and methods help mainly on classification, so they cannot meet the diverse content moderation requirements. Also, without accounting for the moderation process, incorrect moderation explanations are likely produced.

\noindent \textbf{Multimodal Large Language Models.} \ 
LLMs such as Llama 2~\cite{touvron2023llama}, Mixtral~\cite{jiang2024mixtral}, ChatGPT~\cite{ChatGPT}, and GPT4~\cite{achiam2023gpt} demonstrate great achievements in recent years, by scaling (e.g., model/data size) a pre-trained language model to improve its capacity on downstream tasks. To extend the model abilities to visual tasks, many works combine visual and linguistic models for cross-modal comprehension, including well-developed closed-source models such as Gemini~\cite{team2023gemini} and GPT-4V~\cite{achiam2023gpt} and open-source models such as~\cite{liu2024visual, bai2023qwen, ye2023mplug2, dong2024internlm, wang2023cogvlm}. Recently,~\cite{jha2024memeguardllmvlmbasedframework} explored a meme (images with text caption that is typically humorous or sarcastic) intervention method with an MLLM-based pipeline for understanding and intervening meme. Yet, it relies on text recognition and cannot provide results on moderation classifications and explanations following the established rules. Considering the strong scalability of open-source models, we build ICM-Assistants on them, e.g., LLaVA~\cite{liu2024visual}, mPLUG-Owl2~\cite{ye2023mplug2}, and Qwen-VL~\cite{bai2023qwen}.

\section{Methodology}
\subsection{Problem Statement}
As an image moderator under certain moderation rules, given an input image, existing neural-network-based methods normally regard such an ICM as a simple classification task. It can be formulated as maximizing a posteriori of the moderation classification result conditioned on the input and network parameters $\phi_{r}$ that follow moderation rules $r$:
\begin{equation}
\phi^* = \mathop{\arg\max}\limits_{\phi} p( \mathcal{C} | I, \phi_{r} ),
\label{ps1}
\end{equation}
where $I$ is the input image and $\mathcal{C}$ indicates the classification result on a moderation term, following pre-defined rules $r$.
However, for explainable moderation and incorporating the explanation process in the moderation classification results, our ICM-Assistant outputs moderation results additionally with a moderation explanation $\mathcal{E}$, by optimizing
\begin{equation}
\Theta^* = \mathop{\arg\max}\limits_{\Theta} p( \mathcal{E} | I, \Theta_{r} ),
\label{ps2}
\end{equation}
where $\Theta_{r}$ is the parameters of the ICM-Assistant with rules $r$ and $\mathcal{E}$ is the moderation explanation including \mbox{$<\mathcal{D}_{e.e.}, \mathcal{D}_{e.i.}, \mathcal{E}_{e.i.}>$}, in which $\mathcal{D}_{e.e.}$, $\mathcal{E}_{e.i}$, and $\mathcal{D}_{e.i.}$ denote the enriched explicit description, explicit-implicit description, and explicit-implicit explanation, respectively.

Based on this proposed model, we here cast the rule-based moderation task as a general pipeline, from data generation to MLLM instruction-tuning. As illustrated in Fig.~\ref{fig:fig-framework}, given sets of rules on a moderation term (e.g., ``sexy''), we first initialize the explicit descriptions based on the decomposed rules and attributes product. Then, with such descriptions and moderation labels based on the rules, we enrich the explicit descriptions with MLLMs as moderation explanations and extract moderation Q-A with LLMs for a corresponding ICM-Instruct dataset. Finally, we instruction-tune the generated datasets and optionally the existing datasets for the corresponding ICM-Assistant models. 

\begin{figure*}[t]
    \centering
\includegraphics[width=1.\linewidth]{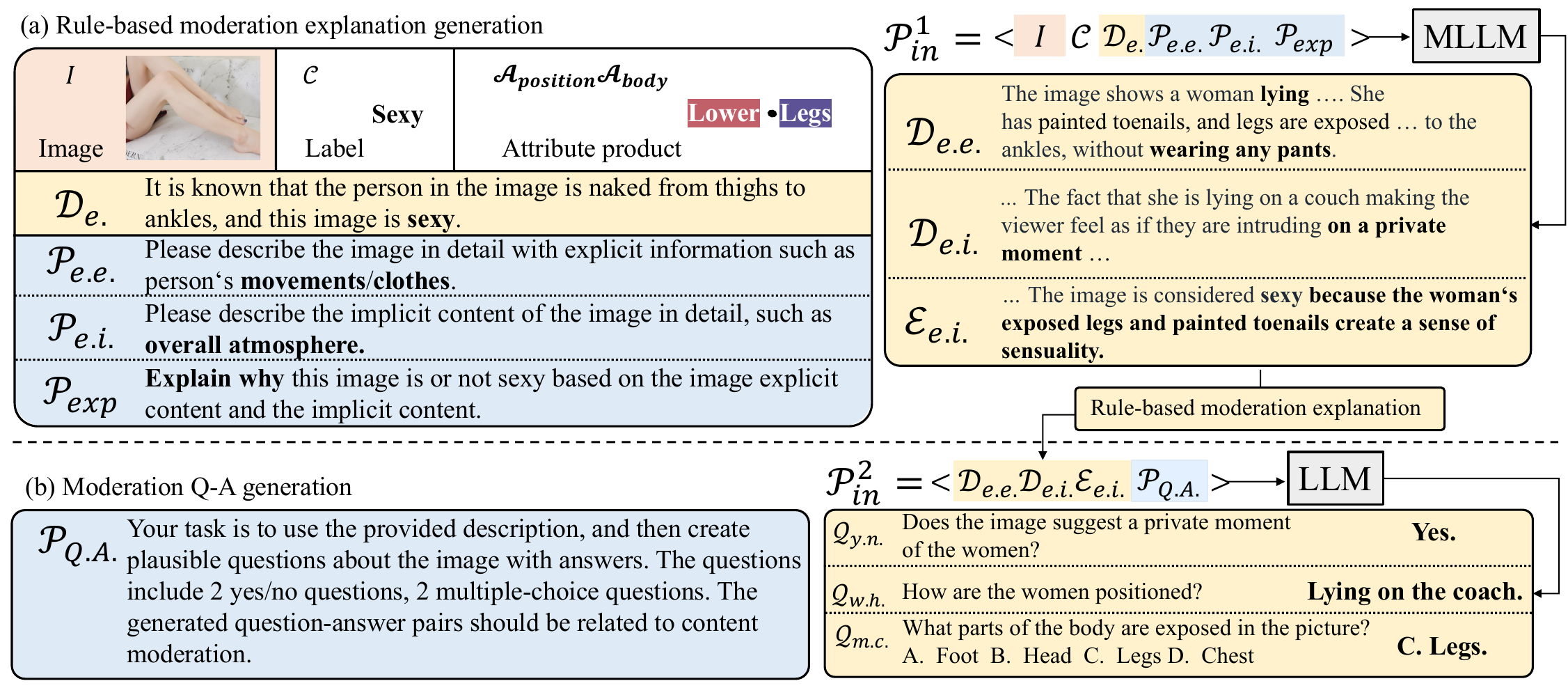}
\caption{Pipeline for generating moderation explanations and Q-A pairs, from one attribute products ``lower-leg''.}
    \label{fig:fig-pipeline}
\end{figure*}
\subsection{Rule Decomposition and Explication Descriptions}
\label{sec:3-1}

Observing the human moderators, they usually achieve proper moderation by first describing the images and then locating regions/semantics that violate the moderation rules. To reduce the workload and achieve flexible moderation, we start with a rule decomposition pipeline to split the rule items into short annotations for collecting images and writing explicit descriptions (one sentence or phrase related to the moderation rules) by moderators. 
Based on our observations, the rules defined for a moderation \textbf{term} (e.g., ``sexy'') can be decomposed into multiple orthogonal basis vectors, terms as \textbf{attributes}. The values assigned to each orthogonal attribute are deliberately chosen to be distinctive in semantics (e.g., lower, middle, and higher in position), such that all the rule items can be represented as products of such attributes are distinct, termed \textbf{attributes products}.

To better explain, we use ``sexy'' as an example moderation term~\cite{Content_policies_for_Google_Search, Content_policies_for_Facebook} to illustrate (Fig.~\ref{fig:fig-framework} (top)):
(1) images with sexy attraction content (e.g., a person's actions and clothing) are considered violations. 
Our method thereby decomposes the moderation rule items for ``sexy'' on the aspects of body bareness, actions of people, and camera focus position on the image. We then retrieve attribute products to summarize possible violations in the rule items, e.g, ``lower leg'' and ``upper take off.'' 
Based on the attribute products, a dataset with real images is downloaded from the Internet by searching them on multiple search engines, including Google, Baidu, and Bing. A set of explicit descriptions paired with moderation classification results for each attribute product is also prepared by human moderators as well (see Fig.~\ref{fig:fig-framework} (top right) and \textit{Supp. A}; in our implementation, there are 35 short sentences in total for the moderation term ``sexy,'' following an example rule).

\subsection{Rule-based Moderation Explanation Generation}
\label{sec:3-2}
With initial explicit descriptions and the paired images set prepared, we generate the {moderation explanation} for each image step by step, by enriching the shared explicit description with the assistance of current pre-trained MLLMs~\cite{chen2023sharegpt4v, wang2023cogvlm}, (see Fig.~\ref{fig:fig-pipeline} (a)): 
(1) We create a set of description-enrich prompts to enrich and include implicit descriptions from given explicit descriptions: {enrich-explicit prompt}: describe the explicit content of the image in detail; {explicit-implicit prompt}: describe the implicit content (e.g., overall atmosphere) of the image; {explanation prompt}: explain whether the image is sexy or not, based on the explicit and implicit contents). (see the details of prompts in \textit{Supp. B})
(2) We use the above prompts as the input of pre-trained MLLMs and a tuple $<\mathcal{D}_{e.e.}, \mathcal{D}_{e.i.}, \mathcal{E}_{e.i.}>$. (enriched-explicit description, explicit-implicit description, explicit-implicit explanation) as moderation explanation $\mathcal{E}$.
(3) We progressively increase the temperature parameters of MLLMs and repeat steps (1) and (2) to produce five samples per image.
Based on images and several shared common explicit descriptions, our method can efficiently extends the amount, length, and variety of the moderation explanation data. 

\subsection{Moderation Q-A Generation}
\label{sec:3-3}
Though the explanation data comes with rich implicit and explicit information inside images, irrelevant content may be introduced in the previous stage. To encourage the model to focus on content directly related to the moderation terms and align it to follow a variety of instructions~\cite{wu2023q, liu2024visual}, we include additional data types, including \textit{Yes/No} questions for binary answers, \textit{What/How} questions for short-sentence answer, and \textit{Multiple-Choice} questions for selected answers. 
With pre-trained LLMs~\cite{touvron2023llama, bai2023qwen}, we thereby generate moderation Q-A from moderation explanation data in the following steps. (see Fig.~\ref{fig:fig-pipeline} (b)):
(1) For each image, we adopt the tuple $<\mathcal{D}_{e.e.}, \mathcal{D}_{e.i.}, \mathcal{E}_{e.i.}>$ (enriched-explicit description, explicit-implicit description, explicit-implicit explanation) as part of the prompt; and
(2) We create {Q-A prompts} with the tuple and append tasks to generate plausible questions about the image with answers. In the prompts, we specify the question types and amounts, then force the output in table layouts (see \textit{Supp. B}).

\subsection{ICM-Instruct dataset}
Due to the huge amount of possible rules and various moderation terms, we use the most common moderation term ``sexy'' \textbf{(S-ICM)} with two different rules \textbf{R1/R2} (\textbf{R1}: a more conservative rule that limits clothing style and skin exposure~\cite{Admin_2022} for children under 13. \textbf{R2}: more open rules without clothing limitations but disallowing body positions with sexual attractions~\cite{motion} for the same age group) as a demonstration of the ICM-Instruct dataset.
In summary, we collect 16K sexy images for training and validation, generate 75K tuples as the moderation explanation data, and create 246K moderation Q-A for the moderation question answering, following the rules R1/R2, termed the \textit{ICM-Instruct} dataset. 

\begin{figure*}[!h]
    \centering
    \includegraphics[width=1.\linewidth]{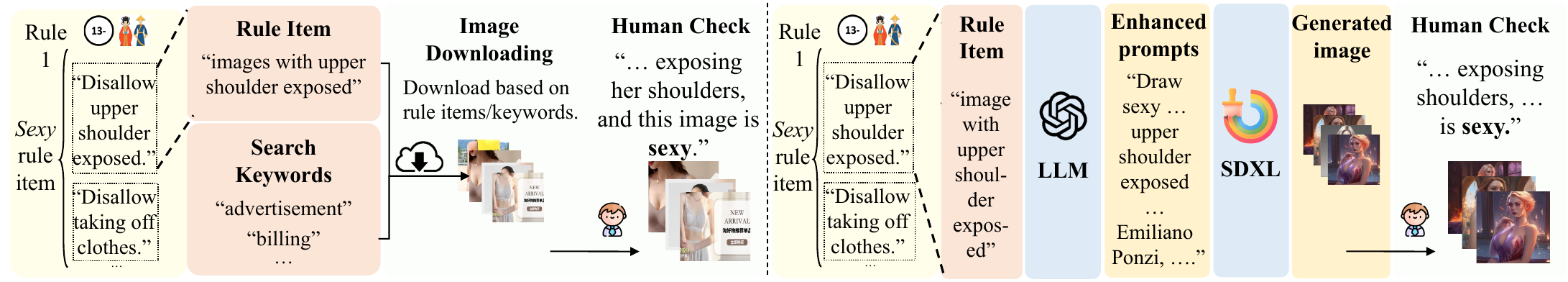}
    \caption{Illustration of the process for building ICM-Test dataset, including ICM-Test-UGC (left) and ICM-Test-AIGC (right).}

    \label{fig:fig-testset}
\end{figure*}

The validation set includes 1,623 images, 1,623 classification labels, 379 moderation explanations from human moderators (a random subset of 1,623 images), and 17,701 moderation Q-A, based on rules R1/R2, termed separately as \textbf{ICM-Val-R1} and \textbf{ICM-Val-R2}.
In addition, there are two test sets adopted in our experiments for further evaluation of the moderation ability for S-ICM, termed \textbf{ICM-Test}, which includes user-generated images and AI-generated images as follows:
(1) \textbf{UGC}: 10,740 user-uploaded advertisement images downloaded from diverse advertisement platforms, where there are 1,443 not-sexy images, labeled according to S-ICM R1. (See Fig.~\ref{fig:fig-testset} (a).)
We keep the data ratio in our test set to simulate real-world scenarios;
(2) \textbf{AIGC}: 13,140 images generated by SDXL-1.0~\cite{podell2023sdxlimprovinglatentdiffusion}. There are 1,000 initial SDXL-styled prompts automatically generated according to S-ICM R1. We generate 1,140 not-sexy images and 12,000 sexy images. (see Fig.~\ref{fig:fig-testset} (b))
Human moderators have carefully checked all the above data samples and labels, and there are no privacy issues, as all the exposed real faces are blurred.
Besides the ICM-Instruct for S-ICM, we collect relatively small sets for moderation terms ``horrifying'' and ``gambling'' for experiments (see \textit{Supp. A}).

Compared with the existing datasets (AIIA-PID4~\cite{karavarsamis2013detecting}, Adult content~\cite{connie2017smart}, NudeNet~\cite{NudeNet}, LSPD~\cite{phan2022lspd}, and MemeGuard~\cite{jha2024memeguardllmvlmbasedframework}) on ICM tasks (Table~\ref{tab:tab_dataset_cmp}), our proposed dataset provides a significantly wider range of sub-categories (based on the attribute products) for moderation, a richer data format with moderation explanation and moderation Q-A for instruction tuning, and modalities that include normal images and meme.
\begin{table}[t]
    \centering
    \resizebox{\linewidth}{!}{
    \begin{tabular}{c|ccc|cc|ccc}
    \hline
         Dataset&\multicolumn{3}{c|}{Categories}&\multicolumn{2}{c|}{Modalities} &\multicolumn{3}{c}{Formats}\\
 & & & & I&M & C& E&Q\\\hline 
 AIIA-PID4& \multicolumn{3}{c|}{4}&\checkmark& & \checkmark& &\\
Adult content& \multicolumn{3}{c|}{2}&\checkmark& & \checkmark& &\\
NudeNet& \multicolumn{3}{c|}{2}&\checkmark& & \checkmark& &\\
 LSPD& \multicolumn{3}{c|}{8}&\checkmark& & \checkmark& &\\
 MemeGuard& \multicolumn{3}{c|}{2}& &\checkmark  & \checkmark& \checkmark&\\
 ICM-Instruct (ours) &\multicolumn{3}{c|}{45}&\checkmark&\checkmark  &\checkmark & \checkmark&\checkmark\\\hline
    \end{tabular}
    }
    \caption{Comparing our ICM-Instruct dataset with existing image (and video) content moderation datasets. In modalities, `I' is for image, `V' for video, and `M' for meme (image with test caption covered). In formats, `C' is for moderation classification, `E' for moderation explanation, and `Q' for moderation Q-A.}
    \label{tab:tab_dataset_cmp}
\end{table} 

\subsection{Instruction-tuning strategies}
In general, the training of open-source MLLMs includes two stages, (1) aligning the representation space of the visual backbone and the LLM with million-scale web data. (2) visual instruction tuning with a combination of human-labeled datasets. Considering the scale of the combination of all the datasets, we adopt a general strategy to mix its instruction-tuning datasets with the existing high-level datasets in the second stage, termed~\textbf{MSFT} (mixed SFT), and a faster and more convenient strategy to add a continuous stage only with the ICM-Instruct after the original high-level tuning of MLLM, termed~\textbf{CSFT} (continuous SFT). In our implementation, we train and validate various open-source MLLMs (e.g., LLaVA~\cite{li2024llava} v1.5 (7B) and v1.6 (7B), Qwen-VL~\cite{bai2023qwen} (7B), mPLUG-Owl2~\cite{ye2023mplug2} (7B)) on the ICM-Instruct dataset.

\begin{table}
    \centering
    \begin{tabular}{l|ll}
    \hline
         Model &  ICM-Val-R1&ICM-Val-R2\\\hline 
            LLaVA-v1.5 &   0.794 & 0.680\\
            ICM-LLaVA-v1.5-R1&   \textbf{0.987}$_{+24.3\%}$ &-\\
 ICM-LLaVA-v1.5-R2& -& 0.971$_{+42.8\%}$\\\hline
            LLaVA-v1.6 & 0.763 & 0.691\\
            ICM-LLaVA-v1.6-R1&   0.986$_{+29.2\%}$&-\\
 ICM-LLaVA-v1.6-R2& -& \textbf{0.973}$_{+40.8\%}$\\\hline
            Qwen-VL & 0.598 & 0.664\\
            Qwen-VL-ICM-R1&   0.967$_{+61.7\%}$ &-\\
 Qwen-VL-ICM-R2& -& 0.961$_{+44.7\%}$ \\\hline
            mPLUG-Owl2 & 0.821 & 0.735\\
            mPLUG-Owl2-ICM-R1&  0.984$_{+19.9\%}$ &-\\
 mPLUG-Owl2-ICM-R2& -& 0.962$_{+30.9\%}$\\\hline
    \end{tabular}
\caption{
Results on ICM accuracy. Our ICM-Assistant consistently brings significant improvement on four different MLLMs for both the ICM-Val sets on rules R1 and R2.
}
\label{tab:tab-rule}
\end{table}

\section{Experiment}
In this section, we present the experimental setting and implementation details and study the ICM abilities and ablations of our framework. 

\textbf{Test set.} 
We employ both the ICM-Val and ICM-Test sets as the main test set. Additionally, to examine the generalizability of our model, we include the open-source NudeNet test set~\cite{NudeNet} (we employ 4,658 neutral images and 13,858 porn images, which naturally follow our rule R1) along with our ICM-Test set for classification accuracy.

\textbf{Metrics.} We employ the following three metrics:
(1) \textbf{ICM Accuracy (ICM acc.):} To better analyze the classification results, instead of directly adapting the outputs (i.e., yes or no) from MLLMs, we proposed an ICM-score to reflect the probability for MLLMs to output with an input prompt such as \textit{``Is the image sexy?''}. By calculating the probability as the \textit{softmax} of MLLMs output tokens~\cite{wu2023q}. To calculate ICM-Accuary, we count the ICM-score larger than 0.5 as a positive reply and perform a binary comparison.
(2) \textbf{Moderation explanation quality (MEQ)}: To measure the accuracy of the moderation explanation of MLLMs, we prompt MLLMs to answer the reasons following the classification result, e.g., \textit{``Explain the reason.''}. Inspired by~\cite{fu2023gptscore}, we applied LLMs~\cite{ChatGPT} to compare the reasons and moderation explanations labeled by human moderators according to precision, fluency, and comprehension (see our prompts in \textit{Supp. C}).
(3) \textbf{Moderation question answering accuracy (MQA acc.)}: As a pathway to moderation explanation, we evaluate the models' performance in answering moderation questions.

\begin{figure*}[t]
    \centering
    \includegraphics[width=1.\linewidth]{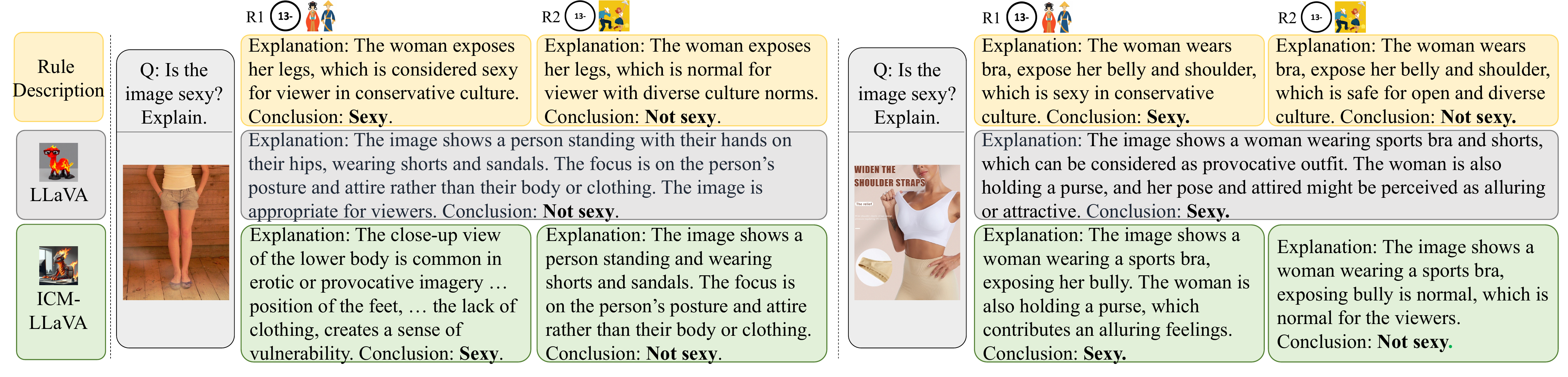}
    \caption{Comparison of ICM results with two different rules, R1 and R2. (See rule differences in Sec: ICM-Instruct dataset.)}
    \label{fig:fig-exp-r}
\end{figure*}

\textbf{Implementation Details.}
All the experiments are conducted on 4x A100 80G GPUs with a default learning rate suggested by their official repositories. To save computing resources, we adopted CSFT in the main experiments. 

\subsection{Results on ICM Ability}
\label{Sec:4-2}

\subsubsection{Results on Flexibility on Rules.}
First, we present the results of the performance enhancement of our ICM-Assistant models trained with datasets constructed on different rules R1 and R2 on S-ICM.
From Table.~\ref{tab:tab-rule}, we can see that with different rules, the accuracy on both ICM-Val-R1 and ICM-Val-R2 shows large improvements on all our models tuned with the corresponding rule-based dataset, with an improvement of 36.8\% on average.
From Fig.~\ref{fig:fig-exp-r}, we can see that with different rules on legs and belly, our models can conclude and explain consistently with the ground-truth rule descriptions provided by human moderators, indicating a large advantage of our models in flexibilities.

\begin{table}
    \centering
    \begin{tabular}{l|ccc}
    \hline
         Model &UGC & AIGC & Pornography \\\hline
            SD Safety Checker &0.040 & 0.307 & 0.589 \\
            NSFW model  &0.089 & 0.679 & 0.637 \\
            NSFW detection&0.020 & 0.231 & 0.481 \\\hline
            ICM-ResNet-152  &0.566 & 0.902 & 0.764 \\
            ICM-DenseNet-201  &0.586 & 0.889 & 0.767 \\
            ICM-CLIP-L/14-336 & 0.772 & 0.942 & 0.817
            \\\hline 
            LLaVA-v1.5 &0.896 & 0.940 & 0.923 \\
            ICM-LLaVA-v1.5&0.958 & 0.970 & \textbf{0.983} \\\hline
            LLaVA-v1.6 &0.929 & 0.957 & 0.923 \\
            ICM-LLaVA-v1.6&\textbf{0.972} & 0.961 & 0.953 \\\hline
            Qwen-VL &0.941 & 0.712 & 0.883 \\
            ICM-Qwen-VL&0.949 & \textbf{0.974} & 0.924 \\\hline
            mPLUG-Owl2 &0.828 & 0.953 & 0.916 \\
            ICM-mPLUG-Owl2&0.937 & 0.965 & 0.906
            \\\hline
    \end{tabular}
\caption{Results on ICM accuracy by traditional methods and various MLLMs on the ICM-Test set.}
\label{tab:tab-1-1}
\end{table}

\subsubsection{Moderation classification.}
\label{sec:4-2-1}
We then experiment with the fundamental ICM classification tasks for the existing ICM methods and also the MLLMs on the Test set (ICM-Test and NudeNet~\cite{NudeNet}).
For quantitative comparison, Table~\ref{tab:tab-1-1} shows the comparison of ICM accuracy of existing open-source ICM methods, similar architectures trained on ICM-Instruct, pre-trained MLLMs, and ICM-Assistant MLLMs on the ICM-Test dataset. 
\textit{First}, since open-source ICM methods~\cite{stable-diffusion-safety-checker, deep-nn-for-nsfw-detection, nsfw_image_detection} are trained on datasets with limited moderation terms (e.g., pornography), it is hard for them to generalize to other related moderation terms (e.g., sexy). Therefore, their results on the UGC and AIGC test sets are unsatisfying. 
\textit{Second}, trained with our ICM-Instruct dataset (as a classification task), several models with similar architectures (e.g., CLIP \cite{radford2021learning}) consistently achieve much better ICM accuracy on all test sets, indicating that the ICM-Instruct dataset helps the traditional models to improve generalization in real-world scenarios. 
\textit{Third}, we also notice that open-source MLLMs~\cite{li2024llava, bai2023qwen, dong2024internlm, ye2023mplug2} have strong fundamental moderation ability on the three ICM-Test sets by outperforming most existing moderation methods. Trained with the full ICM-Instruct dataset, MLLMs consistently achieve much higher accuracy on most test sets. 

\begin{table}
    \centering
    \begin{tabular}{l|ll}
    \hline
         Model & MEQ & MQA acc. \\\hline
            BLIP2 & 0.687& 0.604 \\
            Kosmos-2 & 0.611& 0.138 \\\hline
            LLaVA-v1.5 & 0.696 & 0.801 \\
            ICM-LLaVA-v1.5  & 0.844$_{+21.26\%}$ & \textbf{0.922}$_{+15.11\%}$ \\\hline
            LLaVA-v1.6 & 0.737 & 0.786 \\
            ICM-LLaVA-v1.6  & \textbf{0.846}$_{+14.79\%}$ & 0.922$_{+17.30\%}$\\\hline
            Qwen-VL & 0.541 & 0.768 \\
            ICM-Qwen-VL  & 0.789$_{+45.84\%}$ & 0.910$_{+18.49\%}$ \\\hline
            mPLUG-Owl2 & 0.666 & 0.767 \\
            ICM-mPLUG-Owl2& 0.831$_{+24.77\%}$ & 0.922$_{+20.21\%}$ \\\hline
    \end{tabular}
\caption{
Results on ICM explanation ability on the ICM-Val-R1 set.
}
\label{tab:tab-1}
\end{table}

\begin{figure*}[t]
    \centering
    \includegraphics[width=1.\linewidth]{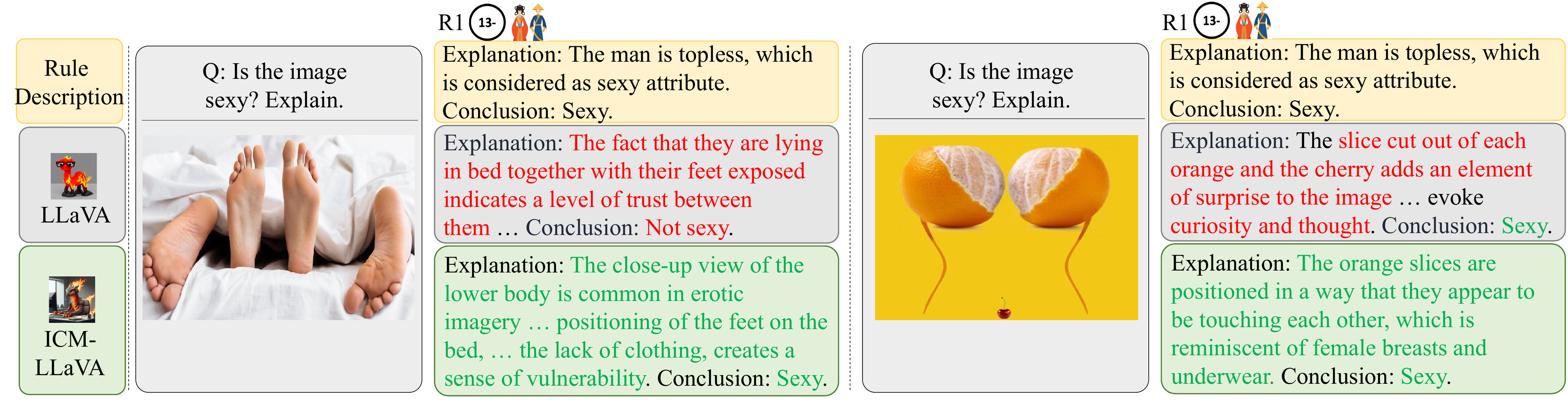}
    \caption{Comparing the moderation quality of ICM-LLaVA-v1.5 with its base model (marked in \textcolor{gray}{gray} and \textcolor{lime}{lime green} separately). \textcolor{green}{Green} texts indicate consistent replies, whereas \textcolor{red}{red} texts indicate inconsistent replies.}
    \label{fig:fig-cmp}
\end{figure*}

\subsubsection{Moderation explanation.}
\label{sec:4-2-2}
Moderation explanation evaluation is based on MEQ and MQA acc. From Table~\ref{tab:tab-1}, we notice the clear advantages of our ICM-Assitant models. Compared with the pre-trained MLLMs, our model achieves higher scores, with 26.6\% higher in MEQ and 17.8\% higher in MQA acc. on average. Fig.~\ref{fig:fig-cmp} shows examples of moderation explanation of LLaVA-v1.5 and our ICM-LLaVA-v1.5. For LLaVA-v1.5, Fig.~\ref{fig:fig-cmp} (left) shows its inconsistency in both classification and moderation explanation results, by missing the possibility of unseen sexual acts. In Fig.~\ref{fig:fig-cmp} (right), though the image is considered to have a sexual act, the shape, appearance, and position of the orange and cherry are missed by LLaVA-v1.5 in deducting the reasons. Our method can provide consistent answers with those from human moderators and explain well. More moderation explanation and Q-A results are presented in \textit{Supp. C}.

\begin{table}
    \centering
    \begin{tabular}{l|ll}

    \hline
      Model   & Horrifying & Gambling \\ \hline
      LLaVA-v1.5   & 0.865 & 0.918 \\
      ICM-LLaVA-v1.5  & \textbf{0.891}$_{+3.01\%}$ & \textbf{0.962}$_{+4.79\%}$ \\ \hline
    \end{tabular}
    \caption{Accuracy of two zero-shot moderation terms.}
    \label{tab:tab-3}
\end{table}
\begin{table}[t]
    \centering
    \begin{tabular}{l|ccc}
    \hline
        Strategies&  ICM acc. &  MEQ & L-B \\\hline
         LLaVA-v1.5&  0.794&  0.696& 54.8\\
          ICM-LLaVA-v1.5-MSFT  &  0.986&  \textbf{0.856}&\textbf{56.0}\\
          ICM-LLaVA-v1.5-CSFT  &  \textbf{0.987} &  0.844& 32.0 \\
          \hline
    \end{tabular}
    \caption{Comparing the performance on different fine-tuning strategies. L-B denotes LLaVA-Bench.}
    \label{tab:ab1}
\end{table}

\subsubsection{Discussion on zero-shot ability.}
\label{sec:4-4}To study the generalizability of ICM-Assistants fine-tuned on S-ICM, we employ test sets for other moderation terms, ``horrifying'' and ``gambling.''
Table~\ref{tab:tab-3} shows that models trained with one single term are also beneficial to other terms, revealing that MLLMs acquire the generalization ability of moderation with the assistance of our ICM-Instruct dataset. It shows the possibility that zero-shot abilities in moderation tasks could emerge with instruction tuning.

\subsection{Ablation Study}
\label{sec:4-3}
LLaVA-v1.5 (7B) is chosen as the baseline for ablations.

\subsubsection{Instruction-tuning Strategies.}
Next, we study two SFT strategies on LLaVA-v1.5: 
CSFT and MSFT have similar performance in ICM acc. Though MSFT has the best explanation quality by mixing both data sources at the final training stage, CSFT consumes less time (6 hours vs 14 hours) with better ICM acc. results. To evaluate generalization ability, we include LLaVA-Bench~\cite{li2024llava}. Column 3 shows that MSFT models demonstrate better generalizability than the pre-trained LLaVA-v1.5, highlighting the potential of our ICM-Instruct dataset for tuning general tasks.

\subsubsection{Components in our pipeline.}
Further, we provide experiments on the choice of MLLM and LLM in the data generation pipelines, as well as the prompts used in generating data with MLLMs. Compared with the baseline ICM-Instruct dataset generated with ShareGPT4V (7B)~\cite{chen2023sharegpt4v} and Llama 2 (70B)~\cite{touvron2023llama}, we choose CogVLM2 (19B)~\cite{wang2023cogvlm} and Qwen2 (72B)~\cite{touvron2023llama} as alternatives. As shown in Table~\ref{tab:tab_ab}, our method achieves consistent performance on each combination, displaying only a small dependency on the choice of the component LLMs and MLLMs in the pipeline.

\begin{table}[t]
    \centering
    \begin{tabular}{cc|ccc}\hline
    MLLM& LLM& ICM acc. &MEQ &MQA acc.\\\hline
 C& Q& 0.977 & 0.811  &0.911 \\
 C& L& 0.978 & 0.802  &0.909 \\
 S& Q& 0.984  & 0.838  &0.915 \\
 S& L& \textbf{0.987}& \textbf{0.844} &\textbf{0.922} \\   \hline
    \end{tabular}
    \caption{Performance of our ICM-Instruct with different choices of MLLMs and LLMs in generating moderation explanation and moderation Q-A. `C' is for CogVLM2, `S' for ShareGPT4V, `Q' for Qwen2, and `L' for Llama 2.}
    \label{tab:tab_ab}
\end{table}

\subsubsection{Data type of Moderation Q-A.}
\label{sec:4-3-3}
Also, we analyze the contribution of data types of moderation Q-A in training the ICM-Assistants in Table~\ref{tab:tab_qa}, with the following findings:
(1) All question types contribute to the ICM acc. (2) \textit{M-C} questions contribute the most to the MQA acc. based on the results of experiments rows 1, 2, and 4. (3) The combination of \textit{Y/N} and \textit{W/H} questions result in the best MEQ, according to experiment row 4. \textit{M-C} questions are not beneficial for generating moderation explanations with rich contents by restricting the answer as a single choice, yet resulting in higher ICM acc. and MQA acc.
\begin{table}[t]
    \centering
    \begin{tabular}{ccc|ccc}
    \hline
   Y/N& W/H& M-C& ICM acc. &  MEQ & MQA acc.\\\hline 
   \checkmark& &  & 0.973&  0.844 &  0.835 \\
   & \checkmark& &0.978  & 0.841  & 0.810 \\
   &  &\checkmark& 0.895 & 0.831  & 0.894 \\
  \checkmark&  \checkmark& & 0.984 & \textbf{0.847}  & 0.830 \\
  \checkmark&   &\checkmark& 0.986 &   0.841& 0.920 \\
    & \checkmark&\checkmark&  0.973&   0.838& 0.915 \\
   \checkmark& \checkmark&\checkmark&  \textbf{0.987} &   0.844&  \textbf{0.922} \\  \hline
    \end{tabular}
    \caption{Performance of our ICM-Instruct with different types of moderation Q-A pairs (Y/N for Yes/No, W/H for What/How, and M-C for Multiple-Choice).}
    \label{tab:tab_qa}
\end{table}

\section{Conclusion}
In this work, we propose a flexible, explainable, and accurate ICM approach by introducing a novel moderation rule-based data generation pipeline, constructing the ICM-Instruct dataset, and creating several instruction-tuned MLLMs as ICM-Assistants to accomplish diverse explainable content moderation tasks. Experimental results show that our approach moderates web-collected images, user-generated advertisements, and AI-generated images with multiple rules. Meanwhile, the zero-shot ability in other moderation terms indicates the possibility of extending the moderation tasks as a general ability for MLLMs. The balance of the dataset between different moderation terms and multi-task learning in instruction tuning will be our future works.

\section{Acknowledgments}
The work is supported by the National Natural Science Foundation of China (No. 62306220) and the RGC/GRF grant 16206524.

\bibliography{aaai25}

\end{document}